# Watershed for Artificial Intelligence: Human Intelligence, Machine Intelligence, and Biological Intelligence

Li Weigang[1], Liriam Enamoto[1], Denise Leyi Li[2], Geraldo Pereira Rocha Filho[1]

[1]Department of Computer Science - University of Brasilia (UnB)
70.910-970 – Brasilia – DF – Brazil

[2]Faculty of Economics, Administration, Accounting and Actuaries - FEA - University of Sao Paulo, Sao Paulo – SP – Brazil

`{weigang,geraldof}@unb.br,{liriam.enamoto,denise.leyi}@gmail.com`

***Abstract.*** *This article reviews the "Once learning" mechanism that was proposed 23 years ago and the subsequent successes of "One-shot learning" in image classification and "You Only Look Once – YOLO" in objective detection. Analyzing the current development of Artificial Intelligence (AI), the proposal is that AI should be clearly divided into the following categories: Artificial Human Intelligence (AHI), Artificial Machine Intelligence (AMI), and Artificial Biological Intelligence (ABI), which will also be the main directions of theory and application development for AI. As a watershed for the branches of AI, some classification standards and methods are discussed: 1) Human-oriented, machine-oriented, and biological-oriented AI R&D; 2) Information input processed by Dimensionality-up or Dimensionality-reduction; 3) The use of one/few or large samples for knowledge learning.*

## 1. Introduction

Artificial Intelligence (AI) has gone through more than sixty years from the conception to the formation of the scientific field [Russell and Norvig, 2002; Luger, 2005; Floreano and Mattiussi, 2008; Vapnik, 2013; Wu Fei, 2020]. Whether it is knowledge engineering based on logical symbols or machine learning (ML) proficient in digital computing, AI science's rapid and continuous development has obtained outstanding achievements. In particular, the extraordinary achievements of deep learning in the fields of natural language processing, image and video processing, and the Internet of Things have accelerated civilized society into the era of AI.

**Some important landmark discoveries in the development of AI will impact the subsequent development of the field**. IBM's Deep Blue Project and Google DeepMind's AlphaGo, and a series of cases that challenge human intelligence, have epoch-making significance in the history of AI development [Campbell et al., 2002; Silver et al., 2016]. This article does not intend to review all processes comprehensively and will focus on the discussion of several relevant academic points. One can see from the in-depth study of the human learning process that the human *a priori* knowledge framework at the beginning of human life is built upon practical knowledge. In particular, the development of computer technology and network technology has made it possible to process and learn massive amounts of information. On this basis, even if a small amount of information is obtained for some special situations, the knowledge will be learned, and solutions will be formed. For example, the parallel processing

mechanism of "Once learning" two-dimensional, holographic information aims to simulate the phenomenon of "Once seen, never forgotten" (in Chinese: 过目不忘) with neural networks [Li Weigang and Silva, 1999]. The "One-shot learning" [Li Fei-fei et al., 2003] method for object categories tries to reproduce the human meta-learning mechanism with one or few examples [Miller et al., 2000]. The "You Only Look Once-YOLO" and "Single shot detector-SSD" proposed novel learning models for object detection, which are widely used [Redmon et al., 2016; Liu et al., 2016].

**Largest technology companies lead R&D of new theories and technologies in AI**. Scholars have gone all the way from neural networks to deep learning in terms of natural language processing. Google DeepMind first launched the AlphaGo series in 2016, using such influential examples to defeat Go masters to push artificial intelligence into society in an all-around way [Silver et al., 2016]. In 2017, the Transformer method was proposed [Vaswani et al., 2017]. Then the BERT model was introduced to learn the whole sentence and perform unsupervised learning of parallel computing to achieve specific downstream applications such as text generation, translation, and analysis and achieve high precision of the results in language processing [Devlin et al., 2018]. This large corps, which is resource-intensive in funds, talents, equipment, and data, engage in scientific and technological breakthroughs and has become a new trend in AI R&D. Especially the advanced research and development of intelligence in the industry and others, the confusion and challenges brought to the scholars, even if most ordinary researchers follow the trend, they are not able to do what they intend to do, and they are miserable.

Under this general trend, **Machine Learning (ML) and Deep Learning (DL) occupy the mainstream of the R&D of AI**. On April 11, 2021, the authors used several keywords to conduct search research on Google search, see Table 1. As a reference, they first chose Facebook for searching. Google showed 24.5 billion searches. Although AI as a subject term should have considerable influence, the number of searches is nearly 680 million, far less than the 2.38 billion of machine learning. The 2.17 billion times of deep learning and even the search results of Facebook are far from an order of magnitude. Mathematics, an old subject that has been formed for nearly a thousand years, can be called the mother of AI, with only 389 million searches, only 3% of Twitter searches. Similarly, from the literature collected on Google Scholar, there are about 3.19 million scientific and technological documents related to AI, 41% of the literature related to Twitter. Moreover, there are more than 5.36 million documents related to machine learning and 4.9 million articles related to deep learning and mathematics separately.

**Table 1. Keyword search results from Google and Google Scholar (April 11, 2021)**

| Site | Keyword | Facebook | Twitter | ML | DL | AI | Mathematics |
|---|---|---|---|---|---|---|---|
| Google | Numbers ($10^6$) | 24,570 180% | 13,640 100% | 2,3801 17% | 2,170 16% | 680 5% | 389 3% |
|  | Time (sec.) | 0.60 | 0.61 | 0.48 | 0.42 | 0.76 | 0.72 |
| Google scholar | Documents ($10^6$) | 6.71 87% | 7.75 100% | 5.36 69% | 4.90 63% | 3.19 41% | 4.90 63% |
|  | Time (sec.) | 0.07 | 0.07 | 0.03 | 0.03 | 0.07 | 0.05 |

The results in Table 1 reflect the current R&D scenario in AI with ML/DL: 1) Excessive and repeated research in ML/DL waste a lot of human and computing resources; 2) Emphasis on the reduction of dimensionality in numerical computing, ignoring the initial purpose of AI, has affected the overall AI R&D and a great part of the researchers.

**AI needs to outline the theoretical system and scientific classification to guide the development**. Although for decades, the theoretical research of AI has made great progress. Some scholars have proposed AI 2.0 [Pan, 2016], and others have proposed strengthening the combination of symbol and data methods, exploring the human thinking process, and developing a new generation of AI [Evans and Grefenstette, 2018]. The "Artificial Intelligence Development Report" [Li et al., 2019] organized by Tsinghua University and others have actively promoted AI development in China. However, in general, there is still a lack of rigorous theoretical reflections to guide the direction of R&D, resulting in uneven development. In particular, the theoretical framework of AI still needs to be strengthened, and programmatic guiding ideologies are proposed. For example, everyone mentions machine learning in general, but specifically, it should be divided into human-like learning, machine learning, and bio-inspired learning. Human-like learning refers to a learning mechanism that changes machine operating conditions, and realizes human-like machines. Its essence is the intelligent realization for humans. Traditional machine learning should specifically refer to the learning mechanism realized by adapting the human learning mechanism to the operating conditions of the machine. Its essence is machine computing.

Based on the "Once learning" mechanism and looking forward to the development trend of multi-modal learning, this article analyzes the current development of AI, especially the existing problems and challenges, and classifies AI into three basic categories: *Artificial Human Intelligence* (AHI) focuses on R&D of making machine similar to human-like intelligence. For example, the Human-like Robot includes Super Human-like Robot (SHLR) to live with human as partners, General Human-like Robot (GHLR) to assist human as domestic worker and others. *Artificial Machine Intelligence* (AMI) focuses on the use of machine scientific computing capabilities, strengthen machine-like intelligence. For example, Machine-like Robot includes Machine arm, Unmanned Aerial Vehicles (UAV), automated driving system, remote surgery system and others. *Artificial Biological Intelligence* (ABI) focuses on the research of biological intelligence and realizes bio-inspired intelligent activities. For example, Biorobotics includes Spider robot, RoboSwift and others. At the same time, these concepts are also established three macro research directions for AI. All branches should be guided by the AI ethical principles and promote the sustainable R&D through AI education and talent training.

The organization of this article is as follows: the second section reviews three learning models as "Once learning", "One-shot learning" and "You Only Look Once-YOLO," comparing and analyzing their similarities and differences, emphasizing learning by small samples. The third section describes a comparative study between Parallel self-organizing map (PSOM) and BERT technology. Section 4 uses examples to illustrate that some scientific calculations can use common neural networks to obtain the same effects as BERT. The fifth section proposes the classification of AI, including AHI, AMI, and ABI. The sixth section summarizes the classification method such as small samples and large samples, dimensionality reduction, and dimensionality upgrade information processing technology. Finally, section 7 is the conclusion of this article.

## 2. Once Learning, One-shot Learning and You Only Look Once

Humans have the ability to acquire all kinds of information from the natural environment, perform knowledge processing, and sublimate to optimal decision-making. The phenomenon of " Once seen, never forgotten" is to perceive the real world visually, learn and memorize the observed scenes, and make corresponding decisions.

In order to describe this phenomenon and thinking process, the "Once learning" mechanism was proposed [Li Weigang, 1998; Li Weigang and Silva, 1999]. The initial idea is based on self-organizing mapping (SOM) with an unsupervised learning model. Referring to the behavior of human visual knowledge acquisition, reflecting the three characteristics of the "Once seen, never forgotten" action: 1) full-screen information input of two-dimensional images is processed once; 2) unsupervised learning mechanism is established using a few parameters; 3) parallel and synchronous memory processing and knowledge learning for full-screen information are implemented.

Over the past 20 years, with the rapid development of computer technology, especially the increasing promotion of AI technology, many exciting research results have been achieved. There have been some independent repeated discoveries related to researches similar to the Once Learning mechanism.

In 2003, in computer vision, the "One-shot learning" method was proposed [Li Feifei et al., 2003], imitating children's enlightenment learning, learning one (or a few) pictures at a time, and analyzing the memory in full-screen image features, image classification, etc. [Miller et al., 2000; Li Fei-fei et al., 2006]. As the basic learning mechanism of Meta-learning, the current "Few-shot" learning has become a traditional algorithm in the field of AI.

Coincidentally, in 2016, in computer vision, scholars engaged in object detection proposed the "You Only Look Once - YOLO" and "Single shot detector - SSD" models [Redmon et al., 2016; Liu et al., 2016]. When processing a picture, target detection aims to identify the type of object in the picture and at the same time marks the specific location of the object. Different from the previous Region-based method, these methods demonstrate the practicality of Region-free and have achieved great success.

For the three "Once Learning Mechanisms" mentioned above, although the motivation and purpose of the model discovery are different, the first step of the information input process is all one-time. This kind of learning mechanism should occupy a place in AI. It should be pointed out here that the concept of "Once learning" has a deeper meaning and a broader application:

First, "Once learning" is used to describe the "Once seen, never forgotten" phenomenon of human access to information, which includes the learning modes of "One-shot" and "You only look once".

Second, the concept of "Once learning" was first put forward in 1998 [Li Weigang, 1998], which is unsupervised learning that uses a glimpse of a full page of text, a full page of text at a time, and parallel computing. This is consistent with the later concepts of the Transformer method [Vaswani et al., 2017] and the BERT model [Devlin et al., 2018] in natural language processing.

Third, an article related to "Once learning" was officially published on IJCNN99, introducing the algorithm implemented by this learning mechanism in Parallel Self-Organizing Mapping (PSOM) and its application in radar image processing [Li Weigang and Silva, 1999].

Fourth, the "Once learning" mechanism is not only limited to the planar graph learning of two-dimensional information but can also be extended to one-time learning of multi-dimensional information [Valova et al., 2005]. In actual situations, the main channels for humans to absorb knowledge come from the five senses. Therefore, the one-time learning mechanism can be extended to a multi-modal learning mechanism that imitates the human senses.

Fifth, human really hopes to realize a learning scenario of "Once seen, never forgotten." The existing computing models and capabilities do not yet have the corresponding conditions, and more advanced computing equipment and algorithms need to be explored. The "Once learning" parallel learning mechanism is proposed based on quantum algorithms [Li Weigang, 1998], and many scholars have made useful attempts in this regard [Xie Guangjun et al., 2003; Li Fei et al., 2004; Bhattacharyya and Bhowmick; 2014; Konar et al., 2016; Bhattacharyya and Maulik, 2017; Wiśniewska et al., 2020].

## 3. Comparison between Parallel self-organizing map (PSOM) and BERT

In the literature, there are different approaches for parallel SOM: software-based approach [Rauber et al., 2000]; hardware-based approach using GPU [Gajdoš and Platoš, 2013; Gavval et al., 2019]; batch-based SOM [Xiao et al., 2015]; and the quantum computing-based algorithms listed above.

Parallel SOM (PSOM) was proposed by Li Weigang [1998] as a software-based approach modifying Kohonen's SOM algorithm. The main characteristics of PSOM are:
- The model can learn the input data in two dimensions at once;
- The competitive learning and weight update can be done simultaneously for the entire dataset by using parallelism;
- The simplicity of the model architecture by using classical distance function for competitive learning such as Euclidean distance;
- The model can take full advantage of parallel computing using GPU.
- It has the potential to realize quantum computing, especially the Once learning mechanism conforms to the mode of quantum computing to form calculations, avoiding the repetitive input of large amounts of data.

In 2017, Vaswani and others proposed the Transformer, a model architecture that relies entirely on an attention mechanism to handle the dependencies between input and output. This architecture allows more parallelization to compute the sequence of continuous representation and keep long-term dependencies without using Recurrent Neural Network or Convolutional Neural Network. The Transformer architecture inspired other variant models, such as BERT [Devlin et al., 2018] and GPT-3 [Brown et al., 2020].

Conceptually, we can find three interesting relationships between Transformer BERT and PSOM as described in the following subsections.

## 3.1. Unsupervised learning

Both models use unsupervised learning approaches to learn features from the input data.

In the pre-training phase, BERT [Devlin et al., 2018] was trained using two tasks without labeled data: (i) Masked Language Model and (ii) Next Sentence Prediction. The Masked Language Model randomly masks some of the tokens from the input, and the objective is to predict the original vocabulary ID of the masked word. In the Next Sentence Prediction, sentence *A* and sentence *B* are loaded into BERT, and the model predicts whether sentence *B* is the next sentence of sentence *A*.

PSOM can learn features from high order input data and represent them in a low-dimensional topological map in an unsupervised fashion. The training is done in two phases: (i) competitive phase and (ii) cooperation phase. In the competitive phase, some distance function is used to compute the best-matching neuron (BMN) between the input data and the randomly initialized weights in a competitive manner. In the cooperation phase, the best-matching neuron's weights are adjusted as well as the neighborhood neuron's weights in order to move the BMN closer to the input vector.

## 3.2. Parallel computing

Both models use parallelization to learn features from input data by performing matrices operations. This allows both models to use GPU to take advantage of parallel computing.

BERT uses an attention function that maps a query and a set of key-value pairs of input data to an output [Vaswani et al., 2017]. The result of the attention is obtained by the dot product of query matrix and key matrix divided by the square root of key dimension, and softmax function is applied to obtain the weights on the values. The computation of the attention function is done simultaneously on a set of queries, packed together into a matrix. This structure is called a single attention head. In BERT, the attention function is performed in parallel, creating different versions of queries, keys, and matrix values called multi-head attention.

The structure of PSOM consists of a matrix of presynaptic neurons representing the input data and a matrix of postsynaptic neurons representing the clustered output data. In the competitive phase, the Euclidean distance between the input data and the weight matrix is calculated for the entire dataset simultaneously. In the cooperation phase, the weight update is computed by matrix multiplication of special weight transformation matrix and weight matrix itself in parallel operations. This architecture allows parallelization to compute the best-matching neuron.

## 3.3. Few-shot learning in GPT-3

The PSOM model has been developed using the "Once learning" mechanism, that is, a two-dimensional holographic sample, such as a whole image or a whole document, is input into PSOM to realize learning. The "One-shot learning" with the same mechanism has been successfully accepted by academia, and then it has been extended to the mechanism of "Few-shot learning" [Wang et al., 2020].

Researchers use massive amounts of data to pre-train intelligent systems to form a particular knowledge structure. At this time, "Few-shot learning" can come in handy. Even if some new and small samples are used, the intelligent system can learn by only

"Few-shot" based on a specific knowledge framework. ". The system can still achieve the expected results with high performance. These concepts and models have become the basic algorithms of Meta-Learning and are also applied to OpenAI's GPT-3 nature language model, which has achieved great success in translation, question answering, word prediction, and even writing of news [Brown et al., 2020]. One thing that needs to be mentioned is that the "Once learning" mechanism of the PSOM model also has some abilities that the "One-shot learning" and other methods do not have, such as appreciation and translation of poetry.

## 4. Comparative Study between Neural Networks and BERT

This section uses three types of neural networks: self-organizing map (SOM), convolutional neural network (CNN) [LeCun et al., 1989], and bidirectional encoder representation from the transformer (BERT) to perform classification experiments on two sets of English essay data [Enamoto et al., 2021]. Its purpose is to show that some AI applications can be implemented using both common neural networks and deep learning.

### 4.1 Datasets

Two publicly available datasets are introduced here as computational experimental data.
> **20NG:** is a multi-class benchmark English dataset collected from twenty different newsgroups and extracted from Scikit-learn datasets. It contains 20,000 texts categorized into 20 classes with a maximum of 60 words for each document. In our experiments, we used a subset of 2 classes and 970 texts.
>
> **COVID19:** is a multi-class English dataset collected from Kaggle public repository. It contains English posts about COVID-19 extracted from Twitter on April 2020. A subset of 157 tweets categorized into five classes with a maximum of 40 words was used in the experiments.

### 4.2 Models

Three architectures of Neural Networks are organized for experiments: Self-organization Map (SOM), Convolutional Neural Network (CNN) with two layers, and Bidirectional Encoder Representations from Transformers (BERT).
- SOM - The publicly available library MiniSOM was used to execute Self Organizing Map. MiniSOM does not implement Parallel SOM yet.
- CNN2L – It was implemented using Keras and TensorFlow with two convolutional layers.
- SOM - CNN2L – It was implemented to add SOM as an embedding function and combined with CNN2L.
- BERT – The publicly available library Ktrain was used to accomplish the test by BERT.

We tested the above architectures in four different combinations. First, we executed SOM as an unsupervised learning model to classify the 20NG dataset into two classes and the COVID19 dataset into five classes. Next, we used CNN2L with an one-hot vector to perform the same classification tasks. Then, we added SOM as an input embedding function of the CNN2L model. Finally, we used BERT to repeat the same experiments for both datasets.

We run all experiments using 10 fold cross-validation, except for BERT. Due to free GPU using restrictions on Google Colab, BERT was executed only once. In these experiments, no systematic grid search was performed, and the results were based on the hyperparameter's manual adjustment.

### 4.3 Results and Discussion

The experiments are conducted and with the following results, see Table 2.
- SOM as an unsupervised classifier performs poorly (0.60) for 20NG dataset and very low accuracy (0.20) for COVID19.
- Executing a vanilla CNN2L also gives poor results for 20NG (0.61) and COVID19 (0.42).
- Adding SOM as an embedding function combined with CNN2L gives 0.87 of accuracy for 20NG, very close to BERT (0.90). For COVID19, the accuracy is 0.71, higher than BERT (0.68).

Table 2. Simulation Results Using 3 Kinds of Neural Networks

| Datasets | | | Results (Accuracy) | | | |
|---|---|---|---|---|---|---|
| Name | Size | Classes | SOM (10fold) | CNN2L (10fold) | SOM-CNN2L (10fold) | BERT (1fold) |
| 20NG | 970 | 2 | 0.60 | 0.61 | 0.87 | 0.90 |
| COVID19 | 156 | 5 | 0.20 | 0.42 | 0.71 | 0.68 |

The above experiments show that some of the practical problems does not necessary require complex, large-scale deep learning computing. Simple neural network models, such as SOM-CNN2L, can also achieve the expected results. If human and computing resources are limited, using this type of model will be a reasonable choice. In particular, "One-shot learning" with one sample or a few samples is well adapted in meta-learning. It is similar to the human learning mechanism that uses prior knowledge to learn new knowledge. This learning concept differs from emerging deep learning mechanisms such as BERT [Devlin et al., 2018] as it is faced with a deep learning system involving more than one hundred network layers and hundreds of millions of learning parameters. People will inevitably ask: will human intelligence be realized like this? Does the human brain have such complex computing power? Are there other different types of intelligent systems? This introduces the motivation to categorize AI in the following sections.

## 5. Three Perspectives of AI: AHI, AMI and ABI

This section discusses the necessity of establishing a theoretical system for AI and proposes three basic branches of AI to strengthen the direction of AI R&D.

### 5.1. Strengthen the Theoretical System of AI

As mentioned above, AI has developed for more than 60 years and has obtained outstanding achievements. The promotion and application of AI have embodied the contribution of modern science and technology to human civilization. The author has been engaged in AI teaching for more than 20 years, understands the development process of AI, and deeply appreciates the development potential of AI. However, AI development is uneven, especially theoretical research that cannot keep up with the advances in applied technology. AI needs to strengthen the theoretical system as a discipline, focusing on the following main problems and their solutions:

First, AI is a new discipline formed by the development of multiple disciplines. From its birth to today, there have been continuous doubts about AI. It can be stated that AI has developed in controversy.

Second, precisely because AI is developed from multiple disciplines, as a scientific discipline, it has not been recognized by some authoritative departments and academic organizations, which limits the development of the field to a certain extent.

Third, the complementarity of the Internet of Things and AI, as well as its promotion and application in various fields of society has vigorously promoted the development of AI. Especially the application-oriented AI R&D has been intensely developed in some large multinational companies, such as Google, Facebook, Amazon, and others. As a result, theoretical research cannot keep up with the actual development of AI, which brings confusion to ordinary scientific scholars, especially for graduate students.

Fourth, non-computer-specialized scientific researchers from other fields flock to AI application research and tend to be mistaken. They often incorrectly understand machine learning as the whole of AI and are overly biased towards machine learning. This fact will result in the proliferation and disorder of AI R&D, ignoring the original intention of AI and its applications.

Fifth, even some scholars in the computer sciences often prefer research directions that are easy to succeed and be published, and then abuse AI technology, stagnating or even losing their way in the theoretical research of AI.

To this end, this article aims to start from the basic theory system of AI and classify the concepts of the branches. We point out that AI should be clearly divided into three research branches: Artificial Human Intelligence (AHI), Artificial Machine Intelligence (AMI), and Artificial Biological Intelligence (ABI). Specific introductions are given in the following subsections.

**5.2. Artificial Human Intelligence – AHI**

Artificial Human Intelligence (AHI) focuses on the development of human-like robot, human-like learning, and other human-oriented AI. Human-like robot will be the mainstream of AHI. Initially, AHI is proposed to include the following sub-branches:

Human-like Robot, including Super Human-like Robot (SHLR) to live with human as partners, General Human-like Robot (GHLR) to assist human as domestic worker etc., and other related studies.

Human-like Learning theory and technology including Meta-Learning, Once learning, and other new learning approaches.

Quantum computing related studies in AHI.

Human brain science and engineering.

Knowledge engineering, the combination of knowledge engineering and numerical computing.

Research on theory and application of AHI, including wearable smart devices and others.

### 5.3. Artificial Machine Intelligence (AMI)

Artificial Machine Intelligence (AMI) focuses on the development of machine computing capabilities to strengthen machine learning, machine-like robot and other machine-oriented AI. AMI is proposed to include the following sub-branches:

Machine learning, including neural networks, deep learning, and other numerical computing.

Machine-like Robot, including Machine arm, Unmanned Aerial Vehicles (UAV), automated driving system, remote surgery system and others.

Engineering related to implement AMI, such as electronics, machinery, materials, equipment, and computers.

Research on theory and application of AMI, Smart city, Intelligent Transport System, and others.

### 5.4. Artificial Biological Intelligence (ABI)

Artificial Biological Intelligence (ABI) focuses on the realization of bio-inspired learning, Biorobotics and other biological oriented AI under the guidance of AI ethical principles. ABI is proposed to include the following sub-branches:

Bio-inspired Learning and others.

Biorobotics, including RoboSwift, Spider, biomedical engineering, cybernetics, and others.

Genetic Programming, Swarm Intelligence, and others traditional topics.

Research on theory and application of ABI.

## 6. Watershed of AI: Research Subject, Sample Size, and Dimensionality

This section briefly introduces the traditional AI classification and proposes new classification standards and methods.

### 6.1. Traditional AI Classification

The classification of AI is basically derived from some textbooks [Luger, 2005; Russell and Norvig, 2020]. For example, Russell and Norvig's classifies the AI based on a two-dimensional combination of design thinking/action and human/rational classification: Acting humanly: The Turing Test approach; Thinking humanly: The cognitive modeling approach; Thinking rationally: The "laws of thought" approach; Acting rationally: The rational agent approach. This classification based on the realization mode of AI is an effective guide to the development of the field, but it is necessary to further establish the theoretical system of AI.

Another popular AI classification considers Artificial Narrow Intelligence (ANI); Artificial General Intelligence (AGI); Artificial Super Intelligence (ASI) [Gottfredson, 1997; Bostrom, 2006]. This classification is based on an overall intelligence characteristic, which has a positive relevance for the popularization and promotion of AI, but it is not a classification of discipline branches. At the same time, the classification does not have an obvious distinction between boundaries and methods.

The traditional classification of AI is mainly based on the following aspects: 1) The research fields of AI mainly include: robots, unmanned machines; natural language processing; image, audio, and other signal processing; Internet of things intelligence; automatic driving, etc. 2) AI technology, mainly including: search; knowledge expression and logical reasoning; expert system; machine learning/deep learning; data mining; pattern recognition, etc. 3) The application fields of AI mainly include: smart transportation; smart city; e-commerce; smart recommendation; social network; and data science, etc. 4) The research objects of AI mainly include knowledge engineering of logical symbols; machine learning based on digital computing; and integrated methods of combining the two.

The various concepts and classifications mentioned above play a positive role in the development of AI. However, with the progress of science and technology and AI R&D, new concepts and classification methods need to be explored.

## 6.2. Several Classification Standards and Methods

**Divided by research object: Human-oriented vs. Machine-oriented vs. Biological-oriented**. This classification method is relatively intuitive and is classified based on the research object. Human-based intelligence research methods and technologies are classified as human-like intelligence and human-oriented reasoning, such as knowledge-based reasoning systems. Intelligent research methods and technologies that take machines as the main body are classified as machine intelligence, such as classic machine learning. Intelligence research methods and technologies that take biology as the main body are classified as biological intelligence, such as the science and technology of designing and constructing biological devices and equipment.

**Divided by the manner of input samples: Few samples vs. more samples**. The "One-shot learning" method [Li Fei-fei et al., 2003] uses a system with prior knowledge to learn from a small number of samples to identify new objects. This process embodies the human learning model. This type of learning can belong to human-like intelligence. Most machine learning methods require learning a large amount or even a vast amount of knowledge. These supervised learning methods can be classified as machine intelligence.

**Divided by knowledge processing mode: Dimension Up vs. Dimension Reduction**. Humans and most animals have five senses: touch, hearing, sight, smell, and taste. If the intelligent system accepts multiple senses and processes the information comprehensively, just as humans receive five sense information, it is called MultiModal Learning-MML [Baltrušaitis and Morency, 2018]. At present, the research on multi-modal learning among images, videos, audios, and semantics is most popular.

Traditional intelligent systems generally accept one type of sense and perform information processing, known as Monomodal Learning. In order to enhance the high intelligence of the system, the main way is to increase the dimension of information input, which is called ascending dimension for short. This type of research can be summarized as a branch of human-like intelligence.

On the other hand, because neural networks and even computers are good at processing low-dimensional information in the direction of machine learning, dimensionality reduction operations on data are widespread, such as information through One-Hot encoding and various embedding methods. Embedding operation is the use of information dimensionality reduction operations for the convenience of machine calculations. Related research can be classified as machine intelligence.

Table 3 summarizes several essential aspects of AI classification, including intelligent objects, input dimensions, sample collection, knowledge expression, important branches, and various examples.

Table 3. Summary of the Watershed of Artificial Intelligence

|  | Artificial Intelligence - AI | | |
| --- | --- | --- | --- |
|  | AHI | AMI | ABI |
| AI object | Human-oriented | Machine-oriented | Biological-oriented |
| Input dimension | Dimension Up (MML) | Dimension reduction (Embedding) | - |
| Sample | Few data | Big data | - |
| Knowledge | Symbolic and Logic | Numerical computing | - |
| Learning | Human-like Learning | Machine /Deep learning | Heuristic learning |
| Robots | Human-like Robot: SHLR, GHLR | Machine-like Robot: UAV | Biorobotics: Spider robot, RoboSwift |

## 7. Conclusion

This article starts with a review of the "Once learning" model and lists three specific learning methods that appear in the literature: "Once learning," "One-shot learning," and "You Only Look Once-YOLO." These three methods can be summarized as "Once Learning Mechanism" because the three methods are consistent in the sense of information input and learning samples. Among them, "One-shot learning" is generally accepted by the academic community, and it is further extended as "Few-shot Learning," which has become a basic method of meta-learning.

Four models of the neural network, including SOM, CNN, and BERT are used to perform classification experiments on two English datasets. The results show that complex and large-scale deep learning calculations are not necessarily required for some practical problems. Simple neural network models such as SOM-CNN2L can also achieve the expected goals with limited human and computing resources.

The article analyzes the current status of AI, discusses existing challenges, and points out the necessity of strengthening AI theoretical study by categorize three branches: Artificial Human Intelligence (AHI), Artificial Machine Intelligence (AMI), and Artificial Biological Intelligence (ABI). This classification can be considered as a watershed in the theoretical system of AI R&D. Take robot as an example, in the near future, Human-like Robot will be more important for human being including Super Human-like Robot (SHLR), General Human-like Robot (GHLR) and others. Some standards and methods of classification are also discussed such as research object, scale of input samples and dimension up or reduction. Some standards and methods of classification are also discussed such as research object, scale of input samples and dimension up or reduction. In the future research, it is urgent to discuss the following topics: 1) classification standards; 2) classification methods; 3) development plan for every branch; and 4) AI concept and sub branches extensions.

Finally, in the first author's AI graduate class, a doctoral student who is also the co-author of this article introduced the theory and practice of BERT. At the end of the class, she asked everyone a question for discussion. The Google AI team has continuously developed a series of comprehensive AI theories and systems with its superior workforce, equipment, systems, and data resources and has obtained exciting achievements, far ahead of universities and other academic research institutions. There is almost nothing to contribute in this field for the teachers and students from common universities, even if they follow the trend of AI R&D. This brings great confusion and risks to doctoral students' scientific research topics. This kind of challenge is unprecedented. A student from the Department of Civil Engineering stated that science and technology in civil engineering are in the leading position for academic units. However, in computer science, including AI, the state of the art is led mainly by industry. The development of computer science conforms to Moore's Law. Everything is developed at a fast pace. We conclude this article with this question and leave it for readers to join our reflection.

**Acknowledgements**

This work has been partially supported by the Brazilian National Council for Scientific and Technological Development (CNPq) under the grant number 311441/2017-3. The authors would like to thank the developers of MiniSom, Keras, TensorFlow and Ktrain. The authors would like to thank the valuable suggestions from Marcos Dib, Li Tian Cheng, Peng Wei, Wu Fei, Wu Xing and Zhai Ziyang.

### *O divisor de águas para inteligência artificial: inteligência humana, inteligência de máquina e inteligência biológica*

***Resumo.*** *Este artigo analisa o mecanismo de "Once learning", proposto há 23 anos, e os sucessos subsequentes do "One-shot learning" na classificação de imagens e do "You Only Look Once – YOLO" na detecção objetiva. Considerando o desenvolvimento atual da IA, propõe-se que a IA seja classificada de forma clara nas seguintes categorias: Inteligência Humana Artificial (AHI), Inteligência de Máquina Artificial (AMI) e Inteligência Biológica Artificial (ABI). Estas serão as principais diretrizes da teoria e desenvolvimento de aplicativos para IA. Como um divisor de águas para os ramos da IA, alguns métodos e padrões de classificação são discutidos: 1) A P&D da IA a ser orientada para o homem, para a máquina e para a biologia; 2) O processamento por dimensionalidade-up ou redução da dimensionalidade da entrada de informações; e 3) O uso de uma / poucas ou grandes amostras para a aprendizagem do conhecimento.*

# 人工智能的分水岭：类人智能，机器智能和仿生智能

*摘要.* 人工智能从概念产生到学科形成，已走过六十多年的历程。无论是基于逻辑符号的知识工程还是精于数字计算的机器学习，使得该学科迅速和持续发展，成就卓然。本文回顾 23 年前提出的"一次学习"*(Once Learning)*机制，通过平行神经网络来实现"过目不忘"现象，以及随后"一瞥学习"*(One-shot Learning)*在图像分类和"你仅看一次"*(You only look once)*等模型在目标检测的应用，分析目前学科研发状况和存在问题，特别是工业界的智能研发超前，为学术界带来的困惑和挑战，提出人工智能应明确分为：人工类人智能*(Artificial Human Intelligence - AHI)*、人工机器智能*(Artificial Machine Intelligence - AMI)* 和人工仿生智能 *Artificial Biological Intelligence - ABI)*等概念。明确理论和应用发展的几个主要方向，以便指导学科稳健发展。传统的人工智能是按知识表达方式分类，例如符号表达或数字表达。本文提出实际应用中的若干分类标准和方法，包括 1) 智能主体实现是以人为本、以机器为本或以生物为本; 2) 信息输入考虑升维或降维技术差别; 3) 知识学习的小样本或大样本的学习理念之分，等等都可以作为人工智能学科分支的分水岭。文中对人工智能学科基本理论，提出的不成熟看法，仅是拾遗补缺，以求尽美尽善。